\pdfoutput=1
\documentclass[11pt,a4paper]{article}
\usepackage[hyperref]{conll2018}
\usepackage{times}
\usepackage{latexsym}
\usepackage{graphicx}
\usepackage{amsmath}
\usepackage{url}
\usepackage{xspace,amsmath,amsfonts}
\usepackage{tikz}
\usepackage{pgfplots}
\usepackage{xcolor}

\definecolor{bblue}{HTML}{4F81BD}
\definecolor{rred}{HTML}{C0504D}
\definecolor{ggreen}{HTML}{9BBB59}
\definecolor{ppurple}{HTML}{9F4C7C}
\usepackage{array}
\newcolumntype{L}[1]{>{\raggedright\let\newline\\\arraybackslash\hspace{0pt}}m{#1}}
\newcolumntype{C}[1]{>{\centering\let\newline\\\arraybackslash\hspace{0pt}}m{#1}}
\newcolumntype{R}[1]{>{\raggedleft\let\newline\\\arraybackslash\hspace{0pt}}m{#1}}

\aclfinalcopy 

\newcommand{\seqseq}{\textsc{seq2seq}\xspace}
\newcommand{\metric}{\textsc{DiscoveryScore}\xspace}
\newcommand{\pmbot}{\textsc{ProfileMemory}\xspace}
\newcommand{\epmbot}{\textsc{ProfileMemory}$^+$\xspace}
\title{Aiming to Know You Better Perhaps Makes Me a More Engaging Dialogue Partner}

\author{
    Yury Zemlyanskiy\\
     University of Southern California\\
     Los Angeles, CA 90089 \\
     {\tt yury.zemlyanskiy@usc.edu}
     \And Fei Sha \\
   	Netflix\\
   	Los Angeles, CA 90028 \\
   	{\tt fsha@netflix.com\thanks{\it{On leave from U. of Southern California (feisha@usc.edu)}}}
}

\date{}

\begin{document}

\maketitle

\begin{abstract}
There have been several attempts to define a plausible motivation for a chit-chat dialogue agent that can lead to engaging conversations. In this work, we explore a new direction where the agent specifically focuses on discovering information about its interlocutor. We formalize this approach by defining a quantitative metric. We propose an algorithm for the agent to maximize it. We validate the idea with human evaluation where our system outperforms various baselines. We demonstrate that the metric indeed correlates with the human judgments of engagingness.
\end{abstract}
\section{Introduction}
\label{intro}

There has been a significant progress in creating end-to-end data-driven dialogue systems~\citep{DBLP:conf/emnlp/RitterCD11,DBLP:journals/corr/VinyalsL15,DBLP:conf/aaai/SerbanSLCPCB17, DBLP:conf/acl/ShangLL15,DBLP:conf/naacl/SordoniGABJMNGD15}. The general scheme is to view dialogues as a sequence transduction process.  This process is then modeled with the sequence-to-sequence (\seqseq) neural network~\citep{DBLP:conf/nips/SutskeverVL14} whose parameters are fit on large dialogue corpora such as OpenSubtitles~\citep{tiedemann2009news}. What is especially appealing about these systems is that they do not require hand-crafted rules to generate reasonable responses in the open-domain dialogue (i.e., chit-chat) setting.

An important goal of such systems is to be able to have a meaningful and engaging conversation with a real person. Despite the progress, however, this goal remains elusive --- current systems often generate generic and universally applicable responses (to any questions) such as \textit{``I do not know''}. While such responses are reasonable in isolation, collectively too many of them are  perceived as dull and repetitive~\citep{DBLP:conf/naacl/SordoniGABJMNGD15, DBLP:conf/aaai/SerbanSLCPCB17, DBLP:conf/naacl/LiGBGD16, DBLP:conf/emnlp/LiMRJGG16}.

It remains open what metrics to use to optimize a data-driven model to produce highly engaging dialogues~\citep{DBLP:conf/emnlp/LiuLSNCP16}. \citet{DBLP:conf/emnlp/LiMRJGG16,DBLP:conf/naacl/LiGBGD16} propose to use several heuristic criteria: how easy to answer the utterance with non-generic response, how grammatical the response is, etc. \citet{DBLP:journals/corr/abs-1801-07243} suggests to use pre-defined facts about the conversation agents as the context for the dialogue. Specifically, conditioning on those facts (called ``memories'' in their approach), the dialogue becomes ``personalized'', purposefully coherent and is perceived being more engaging.

In this paper, we investigate a different approach which leverages the following intuition: an engaging dialogue between two agents is a conversation that is focused and intends to \emph{\textbf{discover}} information with the goal of increased understanding of each other. In other words, discovering implies asking engaging and inquisitive questions that are not meant to be answered with dull responses.

\emph{How do we use these intuitions to build engaging dialogue chatbots?} Imagine a dialogue between a chatbot and a human. The human has facts about herself and is willing to share with the chatbot. The chatbot has only a vague idea what those facts might be -- for instance, it knows out of 100 possible ones, 3 of them are true.  The chatbot's initial utterance could be random as it has no knowledge of what the 3 are. However, the chatbot wants to be engaging so it constantly selects utterances so that it can use them to identify those 3 facts. This is in spirit analogous to a (job) interview: the HR representative (i.e., our interviewer ``chatbot'') is trying to figure out the personality characteristics (i.e., ``facts'') of the applicant (i.e., the ``human'' interviewee). A successful interview implies that the HR representative was able to get as much information about the applicant as possible within a limited amount of time, while dull and repetitive questions are avoided at all cost. In other words, the amount of gathered information can be seen as a proxy measure to the engagingness of the dialogue.

We have implemented such an ``interview'' setting to validate our intuitions. First, we have developed a metric called \metric that can measure how much information has been gathered by the chit-chat bot after a dialogue. During a dialogue, we show how this metric can be used to guide the chatbot's generation of responses at its turns --- these responses are selected so that they lead to the highest expected \metric. To identify such responses, the chit-chat chatbot needs to simulate how its human counterpart would react. To this end, we have proposed an improved version of the personalized chatbot~\citep{DBLP:journals/corr/abs-1801-07243} and use it as the chit-chat bot's model of the human.  Finally, we perform human studies on the Amazon Mechanical Turk platform and demonstrate the positive correlation between \metric and the engagingness scores assessed by human evaluators on our chit-chat bot.

The rest of the paper is organized as the follows. We discuss briefly the related work in Section~\ref{related}. In Section~\ref{method}, we then describe various components in our approach: the metric \metric for assessing how engaging a dialogue is, a chatbot model that is used in our study, and a response selection procedure for our chatbot to yield engaging conversations. We report empirical studies in Section~\ref{exps} and conclude in Section~\ref{conclusion}.

\section{Related Work}
\label{related}

One of the biggest challenges for chit-chat bots is the lack of the exact objective for models to optimize. This stands in stark contrast to task-oriented dialogue systems~\citep{DBLP:journals/corr/WenGMRSUVY16, DBLP:journals/corr/SuGMRUVWY16a}.

Several heuristic criteria are proposed in~\citep{DBLP:conf/naacl/LiGBGD16,DBLP:conf/emnlp/LiMRJGG16} as objectives to optimize. \citet{DBLP:conf/starsem/AsgharPJL17} proposes humans-in-the-loop to select the best response out of a few generated candidates. \citet{DBLP:conf/acl/ChengXGLZF18} uses an additional input signal -- the specificity level of a response, which is estimated by certain heuristics at training time and can be varied during evaluation.

Another way to address the lack of the explicit objective function is to predict many possible responses at once. \citet{DBLP:conf/aaai/ZhouLCLCH17} maps the input message to the distribution over intermediate factors, each of which produces a different response. Similarly, \citep{DBLP:conf/acl/ZhaoZE17, DBLP:conf/aaai/ShenSND18, DBLP:journals/corr/abs-1805-12352} use variants of \textit{variational autoencoder}. These approaches are complementary to defining the objective for dialogue models, as an external reward can further guide the response generation and simplify learning such one-to-many mappings.

\citet{DBLP:conf/emnlp/LiuLSNCP16} hypothesizes that creating a perfect metric for automatic evaluation (so it can be used to optimize a dialogue model to be more engaging, at least in principle) is as hard as creating human-like dialogue system itself. The authors also note that some of the common automatic evaluation metrics (of generated texts) like BLEU, METEOR or ROGUE correlate poorly with human judgments of engagingness. \citet{DBLP:conf/acl/LoweNSABP17} suggests a metric ADEM, which is trained to mimic human evaluators. While it's shown to have better correlation with the scores assigned by humans,  it also gives preference to safer and generic responses.

In our work, we propose to measure how much information the chit-chat bot has gathered about its human counterpart as a proxy to the engagingness of the dialogue. To the best of our knowledge, this metric has not been explored actively in the design of chit-chat bots.

To apply the metric to generate engaging utterances, the chit-chat bot needs to have a model of how the human partner will respond to its utterances.  To this end, we have used the chit-chat bot developed in~\citep{DBLP:journals/corr/abs-1801-07243} as a base model and improved upon it. That bot, called \pmbot, has a set of memories (basically, factual sentences) defining its persona and can output personalized utterances using those memories. Note that in \citep{DBLP:journals/corr/abs-1801-07243} \pmbot is used as a chit-chat bot to generate contextualized dialogue (so as to be engaging). In our work, however, we use it and its improved version as a model of how humans might chat.  Our chit-chat bots can be any existing ones (such as a vanilla \seqseq model without persona) or another \pmbot with its own persona that is different from what humans might have. The key difference is that our chit-chat bot generates utterances to elicit human counterparts to reveal about themselves while \pmbot in its original work generates utterances to tell stories about itself.

Similar ideas have been explored in cognitive research. \citet{DBLP:conf/cogsci/RotheLG16} analyzed how people ask questions to elicit information about the world within a \textit{Battleship} game \cite{Gureckis2009}. In particular, they proposed to evaluate questions based on \textit{Expected Information Gain} \cite{Oaksford1994}, which is built on the similar principles as \metric.

\section{Method}
\label{method}

In the following section, we describe in details our approach for designing engaging chit-chat bots. We start by describing the main idea, followed by discussing each component in our approach.

\subsection{Main Idea}

The main idea behind our approach is that the chit-chat bot stays in ``discovery'' mode. Its main goal is to identify key aspects of its human counterpart. Algorithmically, it chooses utterances to elicit responses from the human so that the responses increase its understanding of the human.

More formally, imagine each human is characterized by  a collection of $K$ facts  $F=\{z_1, z_2, \ldots, z_K\}$, where $z_1$ is \emph{I was born in Russia}, $z_2$ is \emph{My favorite vegetable is carrot},  and $z_K$ is \emph{I like to swim}. The chit-chat bot has access to a universal set of all candidate facts $\mathcal{U}$, and $F$ is just a subset of $\mathcal{U}$. However, the bot does not know the precise composition of $F$ at the beginning of the conversation. Its goal is to identify the subset (or to reduce the uncertainty about it).  With a bit abuse of terminology, we call $F$ the personality of the human or the persona.

We denote a dialogue as a sequence of sentences $h_N = [s_1, t_1, ..., s_N, t_N]$ where $s_n$ denotes the sentences by the chit-chat bot and $t_n$ denotes the ones by the human.

\subsection{A Metric for Measuring Engagingness}\label{section:model:metric}

The chit-chat bot assumes that the human's response $t$ is generated probabilistically when it is the human's turn to respond to the chit-chat bot's utterance $s$
\begin{equation}
P(t \mid s, F) = \sum_{z \in \mathcal{F}} P(t \mid s, z) P(z \mid s, F)
\end{equation}
Intuitively, the human first decides on which fact $z$ she plans to use (ie, which information she wants to reveal) and based on the fact and the chit-chat bot's question, she provides an answer.

The goal of the \textit{discovery oriented} chit-chat is to maximize the mutual information between the dialogue and the revealed personality
\begin{equation}
\mathcal{I}(F; h_N) = \mathbb{H}[P(F)] - \mathbb{H}[P(F \mid h_N)]
\end{equation}
where $\mathbb{H}[\cdot]$ stands for the entropy of the distribution.  Maximizing the mutual information is equivalent to minimizing the uncertainty about $F$ after a dialogue. Intuitively, the chit-chat bot aims to discover the maximum amount of knowledge about the human.  We thus term this quantity as the \metric.

For simplicity, we assume a uniform prior on which $F$ is. Thus, the key quantity to compute is the entropy of the posterior probability. We proceed in two steps.

\paragraph{Calculating the posterior probability}
We assume that every human's response $t_n$ is independent from the previous dialogue history, conditioned on the immediately previous message, and chatbot's question $s_n$ is independent unconditionally. Thus, the posterior can be computed recursively:
\begin{equation}
    P(F \mid h_N) \approx P (F \mid h_{N-1}) \sum_{f \in F \subset \mathcal{U}} P(z_N = f \mid  s_N, t_N)
\end{equation}
where $z_N$ is the fact used in the $N$th turn. The ``single-turn'' posterior for the specific fact $f$ is computed as (we have dropped the subscript $N$ to be cleaner)
\begin{equation}
P(z = f \mid s, t) = \frac{P(t \mid s, z = f) P(z = f \mid s)}{\sum\limits_{f' \in \mathcal{U}}P(t \mid s, z = f') P(z = f' \mid s)}
\end{equation}
We will make a further simplifying assumption that $P(z=f' \mid s)$ is uniform\footnote{This is only an approximation:  the human will respond to ``what kind of food do you like?'' with any facts that relate to food but definitely not to geographical locations, sports, etc. However, this assumption is not as damaging as long as $P(t \mid s, z = f)$ is almost zero for the $z$ that $P(z = f \mid s)$ should be ignored -- the multiplication would result in zero anyway. Since $z$ refers to the fact, $P(t|s, z = f )$ being almost zero reduces to suggest that for a response $t$, there are just only a very limited number of $s$ (questions) and facts that can be used to generate that response. For example, a response \textit{``I lived in Russia as a child''} can only be elicited from \textit{``Where did you spend your childhood?''} (as question) and \textit{``I was born in Russia''} (as a fact). For any other question and fact pair (such as \textit{``Where did you spend your childhood?''}, and \textit{``I like apples''}), the response would be unlikely. We believe this is largely due to the experimental/data design that has ensured facts are being largely non-overlapping for each personality and the dialogues are in general centered around the facts. We leave to future work on how to refine this approximation.} and compute the posterior approximately
\begin{equation}
P(z = f \mid s, t) \approx \frac{P(t \mid s, z = f) }{\sum_{f' \in \mathcal{U}}P(t \mid s, z = f') }
\end{equation}
Substituting this into the expression for $P(F \mid h_N)$, we obtain
\begin{equation}
\begin{split}
P(F \mid h_N) & \approx \\
P(F \mid h_{N-1})&  \frac{\sum_{f \in F}P(t_N \mid s_N, z_N = f)}{\sum_{f \in \mathcal{U}}P(t_N \mid s_N, z_N = f)}
\end{split}
\end{equation}
Acute readers might have identified this as a form of Bayesian belief update, incorporating new evidence at time $N$. The likelihood $P(t_N \mid s_N, z_N = f)$ depends on how to model how the human generates responses. It is sufficient to note that this probability can be computed conveniently by personalized chatbot models. We postpone the details to the next section.

\paragraph{Calculating the entropy} We make an assumption that the number of facts $K$ assigned to the human is known in advance. Therefore, we can consider only probabilities $P(F \mid h_N)$, where $F$ is of a particular known size.
\begin{equation}
P(F \mid h_N, |F| = K) = \frac{P(F \mid h_N)}{\sum\limits_{F' \subset \mathcal{U}, |F'| = K}P(F' \mid h_N)}
\end{equation}
The entropy of distribution $P(F \mid h_N, |F| = K)$ can be computed directly by enumerating all possible combinations of $K$ facts.

\subsection{ChatBot Models}

In our work, there are two types of chatbot models. The first one is the chit-chat bot who will respond to messages from the human conversation partner. While we can use any existing chatbot models, the key ingredient to our approach is to respond so that the expected gain of knowledge on the human is increased. However, since the chit-chat bot cannot inquire the human with ``if I answer you this, would I gain knowledge?'', it has to estimate the gain in knowledge from its model of the human. The second type of model addresses the aspect of modeling the human. In particular, among the 3 models described below, all 3 can be used as the chit-chat bot models and only \pmbot and \epmbot can be used as the model of humans\footnote{In our empirical studies, we use \epmbot most of the time as it is more powerful than the other two.}.

\paragraph{\seqseq dialogue model} This basic model maps an input message $t$ to a vector representation using the encoder LSTM layer and uses it as an initial state $h^d_0$ for the decoder LSTM layer. The decoder predicts a response $s$ sequentially, word by word via softmax. Both the encoder and the decoder share the same input embeddings table.

\paragraph{\pmbot model} \pmbot~\citep{DBLP:journals/corr/abs-1801-07243} is built on top of \seqseq and uses exactly the same architecture for the encoder. Additionally, it has a list of memory slots (called \emph{profile memory}) and each slot stores a fact, represented by a sentence. Each fact is encoded into a single vector representation using the weighted average of its word embeddings where the embeddings table is shared with the encoder and the decoder. In this work, we call the profile memory as the personality.

The decoder is an LSTM layer with attention over the encoded memories. In essence, the attention mechanism computes a weight for each fact and a weighted sum of the facts form a context vector. The context vector and the hidden states are combined as inputs to a softmax layer to generate words sequentially. For details, please consult~\citep{DBLP:journals/corr/abs-1801-07243}

\paragraph{\epmbot model}\label{section:model:epmbot}  The \pmbot has a weakness that is especially critical to our intent of using it as a model of the human. It has to apply attention at every step, even when responding to messages which are not relevant to any of the facts. Thus it always reveals something about the personality (unless the attention is uniform, generally hard to achieve in practice). To address this issue, we enhance the model with a \textit{DefaultFact}, which does not correspond to any real sentence. It does have a vector representation (as other facts do) except the representation is learned during the training. An advantage is that the \textit{DefaultFact} allows to efficiently train on the dialogue datasets without profile memories, such as OpenSubtitles -- intuitively it is the bucket for ``all other facts'' that the dialogue does not explicitly refer to.

\subsection{Dialoguing with Intent to Discover}\label{section:method:predict_metric}

As a metric, \metric can only be computed over and assess a finished dialogue. How can we leverage it to encourage the chit-chat bot to be more engaging? In what follows, we describe one of the most important components in our approach.

Instead of using the standard maximum a posterior inference for the typical \seqseq (and its variants) to generate a sentence, we proceed in two steps to identify the best utterance that has the potential to yield high \metric. The first step is to generate a large set of candidate utterances (for example, using beam search). The second step is to re-rank these utterances.  We describe the second step in details as the first step is fairly standard.

At the $N$th turn of the dialogue, the chit-chat bot has access to the dialogue history $h_{N-1}$ and an estimate of the human's personality $P(F \mid h_{N-1})$.  Let $s$ be a sentence from the chatbot's candidate set. Since the bot has a model of the human, it can predict the human's response $t$ as
\begin{equation}
t \sim P(\cdot \mid h_{N-1}, s, F) =  P(\cdot \mid s, F)
\end{equation}
where $F$ is used to instantiate the model's memory/facts/personality -- in other words, we query the model to see what kind of utterances the human might respond with.

The value of a possible response $s$, i.e, the expected \metric assuming $s$ and $t$ completes the dialogue with $h_N = [h_{N-1}, s, t]$, is then given by
\begin{equation}
V(s) = \mathbb{E}_{F\sim P(\cdot \mid h_{N-1})}\mathbb{E}_{t\sim P(\cdot \mid s, F)} \mathcal{I}(F; h_N)
\end{equation}
Note that the first expectation is needed as the bot has uncertainty of what personality the human is. In practice, we compute this for each $s$ from the candidate set by sampling $F$ and $t$. We then select the optimal utterance that maximize the value
\begin{equation}
s_N = \arg\max_s V(s)
\end{equation}

\section{Experiments}
\label{exps}

We evaluate empirically the proposed approach in several aspects. First, we investigate the effectiveness of the proposed \epmbot model. This model is especially used to model human interlocutors so that it can be used by the chit-chat bot to estimate how an utterance could elicit the human partner to reveal key facts about her (cf.~Section~\ref{section:method:predict_metric}). Secondly, we investigate whether the proposed metric \metric correlates with the engagingness score of a dialogue assessed by human evaluators.

\subsection{Evaluating \epmbot}
\label{exps:model_training}

We contrast \epmbot to \seqseq and \pmbot. We show that not only \epmbot is a stronger model for personalized chit-chat but also \epmbot does not reveal its personality easily. Being discreet is a highly desirable property when the model is used to simulate the human participating in the dialogue; when the facts are easily revealed, then the chit-chat bot can use generic or irrelevant questions to identify the personality thus the dialogue does not become engaging. 

\subsubsection{As a stronger personalized chatbot}
\label{section:model_training}

\paragraph{Datasets} We train all three models on the original PersonaChat dataset~\citep{DBLP:journals/corr/abs-1801-07243} and the Year 2009 version of the OpenSubtitles corpus~\citep{tiedemann2009news}. 
The PersonaChat data set, which consists of crowdsourced 9000 dialogues (123,000 message-response pairs in total) between two people with randomly assigned personas/personalities. There are total 1155 personalities and each personality is defined by 3 to 5 memories (facts such as \textit{``I was born in Russia''} or \textit{``I like to swim''}). 968 dialogues are set aside for validation and 1000 for testing. We report the perplexity of our models on this test data set. The OpenSubtitles corpus has 322,000 dialogues (1.2 million message-response pairs). During training, we augment samples from OpenSubtitles with random personas, which forces \epmbot to actively prioritize \textit{DefaultFact} over these fake facts. 

\paragraph{Implementation Details} Similarly to \citep{DBLP:journals/corr/abs-1801-07243}, we use a single layer LSTM for both the encoder and the decoder with hidden size of 1024 for all models. The
word embeddings are of size 300 and are initialized with GloVe word vectors \citep{DBLP:conf/emnlp/PenningtonSM14}. All models are trained for 20 epochs to maximize the likelihood of the data by using SGD with momentum with batch size 128. Learning rate is reduced by a factor of 4 if the validation perplexity has increased compared to the previous epoch. We found that \textit{general} post-attention~\citep{DBLP:conf/emnlp/LuongPM15} over encoded memories gives better performance than pre-attention. Weights for encoding memories are being learned during training and are initialized with 0.01 for the top 100 frequent words, and with 1 for others. We found that this simple initialization procedure outperforms the one suggested in \citep{DBLP:journals/corr/abs-1801-07243}.

\begin{table}[t]
\centering
\begin{tabular}{ccc}
  \textbf{Model} & \textbf{Datasets} & \textbf{Perplexity}  \\
  \hline
  \seqseq & P  & 38.08  \\
  \pmbot & P & 34.54 \\
  \hline
  \seqseq & P  & 31.538  \\
  \seqseq & P+O & 30.022 \\
  \pmbot & P & 28.406  \\
  \pmbot & P+O & 27.373 \\
  \epmbot & P & 28.098  \\
  \epmbot & P+O  & \textbf{26.807}  \\ \hline
\end{tabular}
\caption{\small Perplexity on PersonaChat test dialogues by 3 different models. For datasets, P stands for PersonaChat and O for OpenSubtitles. The first two rows are reported by \citep{DBLP:journals/corr/abs-1801-07243}. The rests are from our implementation.
}
\label{table:ppl}
\end{table}

\paragraph{Results} The perplexity on the test dialogues by all of the models is contrasted in Table~\ref{table:ppl}. The first two rows are previously reported in \citep{DBLP:journals/corr/abs-1801-07243}. The rest results are from models implemented by us.  

Our re-implementation of \seqseq and \pmbot  show better performance than what are reported in \citep{DBLP:journals/corr/abs-1801-07243}, likely due to the difference in the amount of data used for training\footnote{\citet{DBLP:journals/corr/abs-1801-07243} only modeled the second person in dialogues, which reduces the training data by half.} , as well as model architecture (post- instead of pre-attention) and optimization procedure (e.g., SGD vs. ADAM).

Including additional data such as OpenSubtitles, in general, improves performance. Our \epmbot performs better than \pmbot. This is the benefit of having \textit{DefaultFact} (cf.~Section~\ref{section:model:epmbot}) which re-directs the attention by the messages and responses that are not related to the real personality away it. On the other end, in \pmbot, the attention has to select a real personality no matter what the messages or responses are.

\subsubsection{As a discreet chatbot}
\label{exps:single_sentence_evaluation}


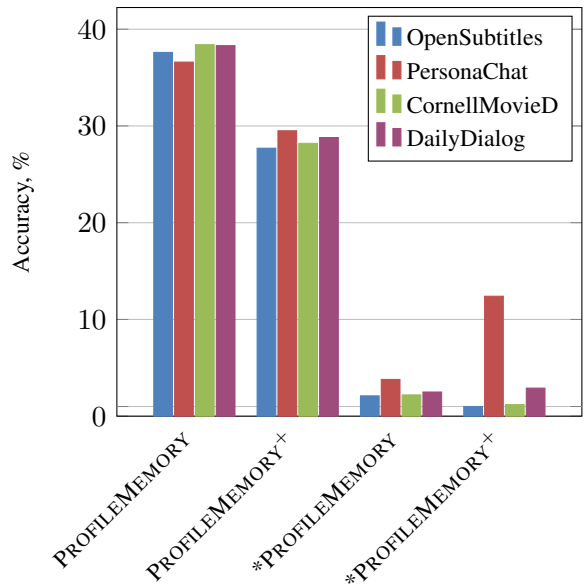
\begin{figure}  
\centering

\noindent\begin{minipage}{\linewidth}
  
\begin{tikzpicture}
        \begin{axis}[
            width  = \textwidth,
            height = 7cm,
            major x tick style = transparent,
            ybar=2*\pgflinewidth,
            bar width=7pt,
            ymajorgrids = true,
            ylabel = {\small Accuracy, \%},
            symbolic x coords={\pmbot,\epmbot, *\pmbot, *\epmbot},
            xtick = data,
            extra y ticks={1},
            extra y tick labels={},
            xticklabel style={rotate=45,anchor=east,font=\small},
            scaled y ticks = false,
            enlarge x limits=0.25,
            ymin=0,
            legend cell align=left,
            legend style={
            }
        ]
            \addplot[style={bblue,fill=bblue,mark=none}]
                coordinates {(\pmbot, 37.6) (\epmbot,27.7) (*\pmbot,2.1) (*\epmbot,1.0)};
    
            \addplot[style={rred,fill=rred,mark=none}]
                 coordinates {(\pmbot,36.6) (\epmbot,29.5) (*\pmbot,3.8) (*\epmbot,12.4)};
    
            \addplot[style={ggreen,fill=ggreen,mark=none}]
                 coordinates {(\pmbot,38.4) (\epmbot,28.2) (*\pmbot,2.2) (*\epmbot,1.2)};
    
            \addplot[style={ppurple,fill=ppurple,mark=none}]
                 coordinates {(\pmbot,38.3) (\epmbot,28.8) (*\pmbot,2.5) (*\epmbot,2.9)};
            
            \legend{\small OpenSubtitles, \small PersonaChat, \small CornellMovieD, \small DailyDialog}

        \end{axis}

\end{tikzpicture}

\end{minipage}

\caption{
\label{fig:retrieval-experiment}
\small Average accuracies of utterances sampled from different corpora (PersonaChat, OpenSubtitles, CornellMovieDialog, DailyDialog) in revealing the personality of the human interlocutor modeled by personalized chatbots (cf.~Section~\ref{exps:single_sentence_evaluation}). \pmbot and \epmbot have been trained on PersonaChat data set, while *\pmbot and *\epmbot have been trained on both PersonaChat and OpenSubtitles.
}
\end{figure}

\begin{table}
\begin{center}
\begin{tabular}{L{7.3cm}}
  \hline
  Top 5 sentences, accuracy 29-32\% \\
  \hline
  \small Tell me about it! What do you do for fun?\\
  \small Nice! what is it that you do? \\
  \small In a cabin, all by myself, hoping my grandkids will visit. any you? \\
  \small You should give it a try! what do you do with your weekends? \\
  \small Wow cool. what do you do in your spare time. i work on art projects. \\
  \hline
  Bottom 5 sentences (sampled), accuracy 0\%  \\
  \hline
  \small How come you were rejected? \\
  \small So he can stay put \\
  \small Spending the night pondering life. \\
  \small Hence the fact that she survived. \\
  \small Maybe she just needs a friend? \\ 
  \hline
\end{tabular}
\end{center}
\caption{\label{table:single-sentence-evaluation-examples}\small Examples of sentences from PersonaChat dataset with the highest and the worst accuracies in revealing a personality of \epmbot model (trained jointly on PersonaChat and OpenSubtitles). We expect human to reply to such utterances with something which will more likely (correspondingly, less likely) reveal her personality. Note, that we don't predict human's personality from the presented utterances alone. Rather, these are considered good (correspondingly, bad) questions to get to know your interlocutor better.}
\end{table}

Since we intend to use a personalized chatbot such as \pmbot and \epmbot as a model of the human interlocutor, we would want the model to behave intelligently: when given an irrelevant message, the model should not reveal its personality. When given a relevant message, the model reveals its personality. We can expect a similar behavior from a real human, which might reply with \textit{``I don't know''} or \textit{``I don't understand''}, when the question is irrelevant to them. In other words, we want to avoid simulating dialogues like this -- Chatbot: \textit{``Hmm... Thank you.''}, Human (Simulation): \textit{``I was born in Russia''}. With this adversity, the chit-chat bot has to ask meaningful and relevant questions if its goal is to discover the personality of the human interlocutor.

\paragraph{Experiment setup} We randomly sample 100 memories/facts (out of total 5709) from the PersonaChat dataset. For simplicity, we assume the model of the human interlocutor has a simple personality, denoted by one of the 100 facts.  We assign this single personality to each of the \pmbot and \epmbot models trained on either the PersonalChat dataset or jointly with the OpenSubtitles dataset. So there are 4 variants in total.

We then construct a simple 2-turn dialogue, where the model is given a probing message and the model responses with a sampled utterance. We use the \metric (cf.~Section~\ref{section:model:metric}) to measure how much the dialogue reveals a personality. We then select the personality that maximizes the revealing. If the selected personality is the true personality, we consider the lead message is able to accurately predict the personality. We then average all probing messages to compute the averaged accuracy.

For probing messages, we use sentences from 4 different datasets - PersonaChat, OpenSubtitles, CornellMovieDialogCorpus~\citep{Danescu-Niculescu-Mizil+Lee:11a} and DailyDialog~\citep{DBLP:conf/ijcnlp/LiSSLCN17}. Our expectation is that an ideal model won't reveal its personality when asked a random question from OpenSubtitles or CornellMovieDialogCorpus, since most of the time it's completely irrelevant lines from a movie script. DailyDialog contains more casual conversations, so some of them we expect to be useful. Of course, the accuracy of random sentences from PersonaChat should be the highest on average, since the corpus was collected with the intent to get to know each other better.

\paragraph{Results} The averaged accuracies from the different corpora are shown in Figure~\ref{fig:retrieval-experiment}.

For \epmbot trained only on PersonaChat data, all types of sentences have similar effectiveness in predicting personality. However, after joint learning with OpenSubtitles, only sentences from PersonaChat (which are most relevant to personalities) are able to predict noticeably accurate than other sentences.

As an illustration, examples of the sentences from PersonaChat with the best and the worst accuracy are presented in the Table~\ref{table:single-sentence-evaluation-examples}, for the \epmbot trained both on PersonaChat and OpenSubtitles.

These findings, together with the superior modeling ability (cf. Section~\ref{section:model_training} and Table~\ref{table:ppl}), have validated the usage of \epmbot trained additionally on OpenSubtitles as a proper model for human interlocutors.

\subsection{Human Evaluation}
\begin{table*}[t!]
\centering
\begin{tabular}{l | lll p{1.7cm}}
  \textbf{ChatBot Model} & \textbf{Fluency} & \textbf{Engagingness} & \textbf{Consistency} & \textbf{Persona Detection, \%} \\
  \hline
  \seqseq & 3.90 (1.24) & 3.52 (1.44) & 3.77 (1.32) & 57.14 \\
  + BeamSearch & 4.25 (1.13) & 3.51 (1.11) & 3.92 (1.27) & 47.41 \\
  + BeamSearch + Re-ranking & 4.64 (0.67) & \textbf{3.92} (1.14) & 4.03 (1.16) & 48.65 \\
  \hline
  \pmbot & 4.13 (1.04) & 3.62 (1.48) & 3.92 (1.29) & 68.57 \\
  + BeamSearch & 4.54 (0.82) & 3.92 (1.09) & 4.28 (1.11) & 60.58 \\
  + BeamSearch + Re-ranking & 4.25 (0.97) & \textbf{4.10} (1.10) & 4.22 (1.06) & 70.00 \\
  \hline
  \epmbot & 4.03 (1.22) & 3.70 (1.22) & 3.79 (1.34) & 78.89 \\
  + BeamSearch & 4.59 (0.84) & 3.73 (1.44) & 4.16 (1.20) & 61.22 \\
  + BeamSearch + Re-ranking & 4.41 (1.05) & \textbf{4.27} (1.07) & 3.99 (1.26) & 69.89 \\ \hline
\end{tabular}
\caption{\small Human evaluation results of various dialogues models. Every model is evaluated by its fluency, engagingness and consistency on a scale from 1 to 5. Persona Detection corresponds to how accurate a human can guess the chatbot's personality thus demonstrating how well a model utilizes the assigned persona (note, it's \textbf{not} related to \metric). Numbers in parenthesis correspond to standard deviation. }
\label{table:human_eval}
\end{table*}
\begin{table*}[t!]
\centering
\begin{tabular}{c|ccc}
Engagingness & Average \metric & \% of questions & Average Length \\
\hline
1 & 2.578 (1.697) & 37.6 & 8.04 \\
2 & 2.776 (1.923) & 48.4 & 8.53 \\
3 & 2.634 (1.837) & 53.0 & 8.29 \\
4 & 2.977 (2.044) & 65.6 & 7.92 \\
5 & 3.196 (1.682) & 55.6 & 8.68 \\
\hline
\end{tabular}
\caption{\label{table:interview-score-vs-engagingness} \small Average \metric over dialogues grouped by corresponding engagingness score in different tiers. Numbers in parenthesis correspond to standard deviation.
}
\end{table*}

\begin{table*}[t]
\centering
\begin{tabular}{l|l}
  \hline
  \small \textbf{ChatBot's Persona} & \small \textbf{Human's Persona} \\
  \hline
  \small I will graduate from college in two years. & \small I play piano. \\
  \small My parents are divorced. & \small I have a ten year old son. \\
  \small I could not live without my dog, Bart. & \small I graduated high school in 1986. \\
  \small I am very popular and have a lot of friends. & \small I hope to become a music teacher after retirement. \\
  \small I am an excellent student. & \\
  \hline
  \hline
  \multicolumn{2}{l}{\small \ [Human]: Hi! How are you doing?} \\
  \multicolumn{2}{l}{\small \epmbot: I am good how are you} \\
  \multicolumn{2}{l}{\small \metric: I am well, thank you. How are you?} \\
  \hline
  \multicolumn{2}{l}{\small \ [Human]: I'm doing okay} \\
  \multicolumn{2}{l}{\small \epmbot: Okay.} \\
  \multicolumn{2}{l}{\small \metric: Ok. What do you do for work?} \\
  \hline
  \multicolumn{2}{l}{\small \ [Human]: I'm working in some local company. How about you?} \\
  \multicolumn{2}{l}{\small \epmbot: I am a stay at home mom} \\
  \multicolumn{2}{l}{\small \metric: I am a college student. Do you have any hobbies?} \\
  \hline
  \multicolumn{2}{l}{\small \ [Human]: I'm a musician, I'm playing piano from time to time.} \\
  \multicolumn{2}{l}{\small \epmbot: What city are you from?} \\
  \multicolumn{2}{l}{\small \metric: That is cool. Do you have any hobbies?} \\
  \hline
  \multicolumn{2}{l}{\small \ [Human]: Not really. But I hope to become a music teacher soon. What are your hobbies?} \\
  \multicolumn{2}{l}{\small \epmbot: What kind of music do you like?} \\
  \multicolumn{2}{l}{\small \metric: I like to spend time with my friends and family. You?} \\
  \hline
\end{tabular}
\caption{\small Examples of the dialogue between human and chatbots: \epmbot model conducts only beam size for best response and \metric additionally uses \metric-based re-ranking.
}
\label{table:dialogue-example}
\end{table*}

We report human evaluations to show that (1) \metric can be used to measure how engaging the conversation is, (2) the dialogue's quality can be increased by choosing a response with the highest expected future \metric (cf.~Section~\ref{section:method:predict_metric}).

\paragraph{Setup} We use ``The Conversational Intelligence Challenge 2''\footnote{\url{http://convai.io}} evaluation procedure provided in ParlAI framework~\citep{DBLP:conf/emnlp/MillerFBBFLPW17}. We use around 200 Amazon Mechanical Turkers for human evaluation. The same procedure is also used in \citep{DBLP:journals/corr/abs-1801-07243}.

During an evaluation round, a Turker is assigned a random persona (with 3-5 profile facts) from the PersonaChat dataset. Each Turker is paired with a chatbot -- we experiment with several models including \seqseq and \pmbot trained on PersonaChat and \epmbot model trained on both  PersonaChat and OpenSubtitles. The chatbot can also adopt personal facts, but only \pmbot and \epmbot are able to utilize it. Every evaluation dialogue has at least 6 turns per participant.

After the dialogue the Turker is asked to evaluate its interlocutor (i.e., the chatbot) by how fluent, engaging and consistent it is on a 1 to 5 scale (5 being the best). Our primary focus is engagingness score and we will show in below that it correlates well with the \metric we proposed. The Turker is also asked to guess the chatbot's persona out of two given persona candidates (each with 3-5 profile facts). This metric is called ``Persona Detection'' and demonstrates how well the model is utilizing the assigned persona. Naturally, we expect \seqseq-based chatbots to have Persona Detection rate around 50\% since they are not using provided persona at all.
Each chatbot model is evaluated on at least 100 dialogues.

\paragraph{Response generation} We experimented with both the greedy decoding (which is default) and the beam search (with 100 beam size) for text generation.

\paragraph{\metric-based re-ranking} As described in Section~\ref{section:method:predict_metric}, we use \metric-based re-ranking to select the response with the intent to discover the personality of the human participant. Concretely, the chatbot is given a set of 30 facts from PersonaChat data set, which does include the true facts. The chatbot is also told that the human has only 3 facts in her personality. This is mainly for computational efficiency.

The re-ranking takes place in two steps. First, the chatbot generates 100 response candidates. For every candidate, it performs 10 simulated dialogues with the \epmbot model as a proxy for the human interlocutor. Finally, it selects the response with the highest expected \metric.

\paragraph{Results} The evaluation results are presented in the Table~\ref{table:human_eval}. The results clearly demonstrate that the \metric-oriented re-ranking makes conversations more engaging for all type of the chatbot models.

When re-ranking is used, many human evaluators provided a feedback stating that the model was acting ``genuinely interested'' and asked a lot of questions. In contrast, modeling without re-ranking had a lower engagingness score precisely because of the lack of questions. 

Persona Detection score indicates that \epmbot is doing a better job in modeling a persona. We also see a decrease in this metric when we combine \epmbot with the re-ranking procedure, which is likely caused by the chatbot asking more questions than revealing itself personality. 

Example dialogues between human and two \epmbot models with and without \metric-based re-ranking are given in the Table~\ref{table:dialogue-example}.

\paragraph{\metric as a proxy for Engagingness} We group all the dialogues between chatbots and humans by the assigned engagingness score and compute the average \metric, average length of utterances and average percentage of generated questions - see Table~\ref{table:interview-score-vs-engagingness}. Interestingly enough, there is no obvious correlation between how engaging the dialogue has been perceived and simple metrics like the length of the response or the number of asked questions. On the other hand, it is strongly correlated with \metric, indicating that it indeed can be used as one of the automatic metrics for dialogues quality.

\section{Conclusion \& Future Work}
\label{conclusion}

We introduce a new metric \metric to assess the engagingness of a dialogue based on the intuition that the more interested the chatbot is in its interlocutor the more engaging the dialog becomes. We propose an improved \epmbot model, which achieves state-of-the-art perplexity results on the PersonaChat dataset. One appealing property of the model is that it doesn't reveal assigned personality upon irrelevant questions. We demonstrate how it can be used to estimate the expected \metric by running simulations with the model as a human substitute. A re-ranking method that uses such estimates allows us to significantly improve the dialogue engagingness score over several baselines, which we demonstrate with human evaluations. 

We hope to continue exploring \metric in more general settings with richer, more complicated personalization or when profile information is not explicitly defined.

\section*{Acknowlegements}
We appreciate the feedback from the reviewers. KG is partially supported by NSF IIS-1514118 and an AWS Machine Learning Research Award. Others are partially supported by USC Graduate Fellowships, NSF IIS-1065243, 1451412, 1513966/1632803/1833137, 1208500, CCF-1139148, a Google Research Award, an Alfred P. Sloan Research Fellowship, gifts from Facebook and Netflix, and ARO\# W911NF-12-1-0241 and W911NF-15-1-0484.

\bibliographystyle{acl_natbib_nourl}



\begin{thebibliography}{31}
\expandafter\ifx\csname natexlab\endcsname\relax\def\natexlab#1{#1}\fi

\bibitem[{Asghar et~al.(2017)Asghar, Poupart, Jiang, and
  Li}]{DBLP:conf/starsem/AsgharPJL17}
Nabiha Asghar, Pascal Poupart, Xin Jiang, and Hang Li. 2017.
\newblock Deep active learning for dialogue generation.
\newblock In \emph{Proceedings of the 6th Joint Conference on Lexical and
  Computational Semantics, *SEM @ACM 2017, Vancouver, Canada, August 3-4,
  2017}, pages 78--83. Association for Computational Linguistics.

\bibitem[{Barzilay and Kan(2017)}]{DBLP:conf/acl/2017-1}
Regina Barzilay and Min{-}Yen Kan, editors. 2017.
\newblock \emph{Proceedings of the 55th Annual Meeting of the Association for
  Computational Linguistics, {ACL} 2017, Vancouver, Canada, July 30 - August 4,
  Volume 1: Long Papers}. Association for Computational Linguistics.

\bibitem[{Cheng et~al.(2018)Cheng, Xu, Guo, Lan, Zhang, and
  Fan}]{DBLP:conf/acl/ChengXGLZF18}
Xueqi Cheng, Jun Xu, Jiafeng Guo, Yanyan Lan, Ruqing Zhang, and Yixing Fan.
  2018.
\newblock Learning to control the specificity in neural response generation.
\newblock In \emph{Proceedings of the 56th Annual Meeting of the Association
  for Computational Linguistics, {ACL} 2018, Melbourne, Australia, July 15-20,
  2018, Volume 1: Long Papers}, pages 1108--1117. Association for Computational
  Linguistics.

\bibitem[{Danescu-Niculescu-Mizil and
  Lee(2011)}]{Danescu-Niculescu-Mizil+Lee:11a}
Cristian Danescu-Niculescu-Mizil and Lillian Lee. 2011.
\newblock Chameleons in imagined conversations: A new approach to understanding
  coordination of linguistic style in dialogs.
\newblock In \emph{Proceedings of the Workshop on Cognitive Modeling and
  Computational Linguistics, ACL 2011}.

\bibitem[{Gu et~al.(2018)Gu, Cho, Ha, and
  Kim}]{DBLP:journals/corr/abs-1805-12352}
Xiaodong Gu, Kyunghyun Cho, JungWoo Ha, and Sunghun Kim. 2018.
\newblock Dialogwae: Multimodal response generation with conditional
  wasserstein auto-encoder.
\newblock \emph{CoRR}, abs/1805.12352.

\bibitem[{Gureckis and Markant(2009)}]{Gureckis2009}
Todd~M Gureckis and Douglas~B. Markant. 2009.
\newblock {Active Learning Strategies in a Spatial Concept Learning Game}.
\newblock \emph{Proceedings of the 31st Annual Conference of the Cognitive
  Science Society}.

\bibitem[{Li et~al.(2016{\natexlab{a}})Li, Galley, Brockett, Gao, and
  Dolan}]{DBLP:conf/naacl/LiGBGD16}
Jiwei Li, Michel Galley, Chris Brockett, Jianfeng Gao, and Bill Dolan.
  2016{\natexlab{a}}.
\newblock A diversity-promoting objective function for neural conversation
  models.
\newblock In \emph{{NAACL} {HLT} 2016, The 2016 Conference of the North
  American Chapter of the Association for Computational Linguistics: Human
  Language Technologies, San Diego California, USA, June 12-17, 2016}, pages
  110--119. The Association for Computational Linguistics.

\bibitem[{Li et~al.(2016{\natexlab{b}})Li, Monroe, Ritter, Jurafsky, Galley,
  and Gao}]{DBLP:conf/emnlp/LiMRJGG16}
Jiwei Li, Will Monroe, Alan Ritter, Dan Jurafsky, Michel Galley, and Jianfeng
  Gao. 2016{\natexlab{b}}.
\newblock Deep reinforcement learning for dialogue generation.
\newblock In  \cite{DBLP:conf/emnlp/2016}, pages 1192--1202.

\bibitem[{Li et~al.(2017)Li, Su, Shen, Li, Cao, and
  Niu}]{DBLP:conf/ijcnlp/LiSSLCN17}
Yanran Li, Hui Su, Xiaoyu Shen, Wenjie Li, Ziqiang Cao, and Shuzi Niu. 2017.
\newblock Dailydialog: {A} manually labelled multi-turn dialogue dataset.
\newblock In \emph{Proceedings of the Eighth International Joint Conference on
  Natural Language Processing, {IJCNLP} 2017, Taipei, Taiwan, November 27 -
  December 1, 2017 - Volume 1: Long Papers}, pages 986--995. Asian Federation
  of Natural Language Processing.

\bibitem[{Liu et~al.(2016)Liu, Lowe, Serban, Noseworthy, Charlin, and
  Pineau}]{DBLP:conf/emnlp/LiuLSNCP16}
Chia{-}Wei Liu, Ryan Lowe, Iulian Serban, Michael Noseworthy, Laurent Charlin,
  and Joelle Pineau. 2016.
\newblock How {NOT} to evaluate your dialogue system: An empirical study of
  unsupervised evaluation metrics for dialogue response generation.
\newblock In  \cite{DBLP:conf/emnlp/2016}, pages 2122--2132.

\bibitem[{Lowe et~al.(2017)Lowe, Noseworthy, Serban, Angelard{-}Gontier,
  Bengio, and Pineau}]{DBLP:conf/acl/LoweNSABP17}
Ryan Lowe, Michael Noseworthy, Iulian~Vlad Serban, Nicolas Angelard{-}Gontier,
  Yoshua Bengio, and Joelle Pineau. 2017.
\newblock Towards an automatic turing test: Learning to evaluate dialogue
  responses.
\newblock In  \cite{DBLP:conf/acl/2017-1}, pages 1116--1126.

\bibitem[{Luong et~al.(2015)Luong, Pham, and
  Manning}]{DBLP:conf/emnlp/LuongPM15}
Thang Luong, Hieu Pham, and Christopher~D. Manning. 2015.
\newblock Effective approaches to attention-based neural machine translation.
\newblock In \emph{Proceedings of the 2015 Conference on Empirical Methods in
  Natural Language Processing, {EMNLP} 2015, Lisbon, Portugal, September 17-21,
  2015}, pages 1412--1421. The Association for Computational Linguistics.

\bibitem[{Miller et~al.(2017)Miller, Feng, Batra, Bordes, Fisch, Lu, Parikh,
  and Weston}]{DBLP:conf/emnlp/MillerFBBFLPW17}
Alexander~H. Miller, Will Feng, Dhruv Batra, Antoine Bordes, Adam Fisch, Jiasen
  Lu, Devi Parikh, and Jason Weston. 2017.
\newblock Parlai: {A} dialog research software platform.
\newblock In \emph{Proceedings of the 2017 Conference on Empirical Methods in
  Natural Language Processing, {EMNLP} 2017, Copenhagen, Denmark, September
  9-11, 2017 - System Demonstrations}, pages 79--84. Association for
  Computational Linguistics.

\bibitem[{Oaksford and Chater(1994)}]{Oaksford1994}
Mike Oaksford and Nick Chater. 1994.
\newblock {A rational analysis of the selection task as optimal data
  selection.}
\newblock \emph{Psychological Review}, 101(4):608--631.

\bibitem[{Pennington et~al.(2014)Pennington, Socher, and
  Manning}]{DBLP:conf/emnlp/PenningtonSM14}
Jeffrey Pennington, Richard Socher, and Christopher~D. Manning. 2014.
\newblock Glove: Global vectors for word representation.
\newblock In \emph{Proceedings of the 2014 Conference on Empirical Methods in
  Natural Language Processing, {EMNLP} 2014, October 25-29, 2014, Doha, Qatar,
  {A} meeting of SIGDAT, a Special Interest Group of the {ACL}}, pages
  1532--1543. {ACL}.

\bibitem[{Ritter et~al.(2011)Ritter, Cherry, and
  Dolan}]{DBLP:conf/emnlp/RitterCD11}
Alan Ritter, Colin Cherry, and William~B. Dolan. 2011.
\newblock Data-driven response generation in social media.
\newblock In \emph{Proceedings of the 2011 Conference on Empirical Methods in
  Natural Language Processing, {EMNLP} 2011, 27-31 July 2011, John McIntyre
  Conference Centre, Edinburgh, UK, {A} meeting of SIGDAT, a Special Interest
  Group of the {ACL}}, pages 583--593. {ACL}.

\bibitem[{Rothe et~al.(2016)Rothe, Lake, and
  Gureckis}]{DBLP:conf/cogsci/RotheLG16}
Anselm Rothe, Brenden~M. Lake, and Todd~M. Gureckis. 2016.
\newblock Asking and evaluating natural language questions.
\newblock In \emph{Proceedings of the 38th Annual Meeting of the Cognitive
  Science Society, Recogbizing and Representing Events, CogSci 2016,
  Philadelphia, PA, USA, August 10-13, 2016}. cognitivesciencesociety.org.

\bibitem[{Serban et~al.(2017)Serban, Sordoni, Lowe, Charlin, Pineau, Courville,
  and Bengio}]{DBLP:conf/aaai/SerbanSLCPCB17}
Iulian~Vlad Serban, Alessandro Sordoni, Ryan Lowe, Laurent Charlin, Joelle
  Pineau, Aaron~C. Courville, and Yoshua Bengio. 2017.
\newblock A hierarchical latent variable encoder-decoder model for generating
  dialogues.
\newblock In  \cite{DBLP:conf/aaai/2017}, pages 3295--3301.

\bibitem[{Shang et~al.(2015)Shang, Lu, and Li}]{DBLP:conf/acl/ShangLL15}
Lifeng Shang, Zhengdong Lu, and Hang Li. 2015.
\newblock Neural responding machine for short-text conversation.
\newblock In \emph{Proceedings of the 53rd Annual Meeting of the Association
  for Computational Linguistics and the 7th International Joint Conference on
  Natural Language Processing of the Asian Federation of Natural Language
  Processing, {ACL} 2015, July 26-31, 2015, Beijing, China, Volume 1: Long
  Papers}, pages 1577--1586. The Association for Computer Linguistics.

\bibitem[{Shen et~al.(2018)Shen, Su, Niu, and
  Demberg}]{DBLP:conf/aaai/ShenSND18}
Xiaoyu Shen, Hui Su, Shuzi Niu, and Vera Demberg. 2018.
\newblock Improving variational encoder-decoders in dialogue generation.
\newblock In \emph{Proceedings of the Thirty-Second {AAAI} Conference on
  Artificial Intelligence, New Orleans, Louisiana, USA, February 2-7, 2018}.
  {AAAI} Press.

\bibitem[{Singh and Markovitch(2017)}]{DBLP:conf/aaai/2017}
Satinder~P. Singh and Shaul Markovitch, editors. 2017.
\newblock \emph{Proceedings of the Thirty-First {AAAI} Conference on Artificial
  Intelligence, February 4-9, 2017, San Francisco, California, {USA}}. {AAAI}
  Press.

\bibitem[{Sordoni et~al.(2015)Sordoni, Galley, Auli, Brockett, Ji, Mitchell,
  Nie, Gao, and Dolan}]{DBLP:conf/naacl/SordoniGABJMNGD15}
Alessandro Sordoni, Michel Galley, Michael Auli, Chris Brockett, Yangfeng Ji,
  Margaret Mitchell, Jian{-}Yun Nie, Jianfeng Gao, and Bill Dolan. 2015.
\newblock A neural network approach to context-sensitive generation of
  conversational responses.
\newblock In \emph{{NAACL} {HLT} 2015, The 2015 Conference of the North
  American Chapter of the Association for Computational Linguistics: Human
  Language Technologies, Denver, Colorado, USA, May 31 - June 5, 2015}, pages
  196--205. The Association for Computational Linguistics.

\bibitem[{Su et~al.(2016{\natexlab{a}})Su, Carreras, and
  Duh}]{DBLP:conf/emnlp/2016}
Jian Su, Xavier Carreras, and Kevin Duh, editors. 2016{\natexlab{a}}.
\newblock \emph{Proceedings of the 2016 Conference on Empirical Methods in
  Natural Language Processing, {EMNLP} 2016, Austin, Texas, USA, November 1-4,
  2016}. The Association for Computational Linguistics.

\bibitem[{Su et~al.(2016{\natexlab{b}})Su, Gasic, Mrksic, Rojas{-}Barahona,
  Ultes, Vandyke, Wen, and Young}]{DBLP:journals/corr/SuGMRUVWY16a}
Pei{-}Hao Su, Milica Gasic, Nikola Mrksic, Lina~Maria Rojas{-}Barahona, Stefan
  Ultes, David Vandyke, Tsung{-}Hsien Wen, and Steve~J. Young.
  2016{\natexlab{b}}.
\newblock Continuously learning neural dialogue management.
\newblock \emph{CoRR}, abs/1606.02689.

\bibitem[{Sutskever et~al.(2014)Sutskever, Vinyals, and
  Le}]{DBLP:conf/nips/SutskeverVL14}
Ilya Sutskever, Oriol Vinyals, and Quoc~V. Le. 2014.
\newblock Sequence to sequence learning with neural networks.
\newblock In \emph{Advances in Neural Information Processing Systems 27: Annual
  Conference on Neural Information Processing Systems 2014, December 8-13 2014,
  Montreal, Quebec, Canada}, pages 3104--3112.

\bibitem[{Tiedemann(2009)}]{tiedemann2009news}
J{\"o}rg Tiedemann. 2009.
\newblock News from {OPUS} -- a collection of multilingual parallel corpora
  with tools and interfaces.
\newblock In \emph{Proceedings of Recent Advances in Natural Language
  Processing (RANLP)}, volume~5, pages 237--248.

\bibitem[{Vinyals and Le(2015)}]{DBLP:journals/corr/VinyalsL15}
Oriol Vinyals and Quoc~V. Le. 2015.
\newblock A neural conversational model.
\newblock \emph{CoRR}, abs/1506.05869.

\bibitem[{Wen et~al.(2016)Wen, Gasic, Mrksic, Rojas{-}Barahona, Su, Ultes,
  Vandyke, and Young}]{DBLP:journals/corr/WenGMRSUVY16}
Tsung{-}Hsien Wen, Milica Gasic, Nikola Mrksic, Lina~Maria Rojas{-}Barahona,
  Pei{-}Hao Su, Stefan Ultes, David Vandyke, and Steve~J. Young. 2016.
\newblock A network-based end-to-end trainable task-oriented dialogue system.
\newblock \emph{CoRR}, abs/1604.04562.

\bibitem[{Zhang et~al.(2018)Zhang, Dinan, Urbanek, Szlam, Kiela, and
  Weston}]{DBLP:journals/corr/abs-1801-07243}
Saizheng Zhang, Emily Dinan, Jack Urbanek, Arthur Szlam, Douwe Kiela, and Jason
  Weston. 2018.
\newblock Personalizing dialogue agents: {I} have a dog, do you have pets too?
\newblock \emph{CoRR}, abs/1801.07243.

\bibitem[{Zhao et~al.(2017)Zhao, Zhao, and
  Esk{\'{e}}nazi}]{DBLP:conf/acl/ZhaoZE17}
Tiancheng Zhao, Ran Zhao, and Maxine Esk{\'{e}}nazi. 2017.
\newblock Learning discourse-level diversity for neural dialog models using
  conditional variational autoencoders.
\newblock In  \cite{DBLP:conf/acl/2017-1}, pages 654--664.

\bibitem[{Zhou et~al.(2017)Zhou, Luo, Cao, Lin, Chen, and
  He}]{DBLP:conf/aaai/ZhouLCLCH17}
Ganbin Zhou, Ping Luo, Rongyu Cao, Fen Lin, Bo~Chen, and Qing He. 2017.
\newblock Mechanism-aware neural machine for dialogue response generation.
\newblock In  \cite{DBLP:conf/aaai/2017}, pages 3400--3407.

\end{thebibliography}

\end{document}